%% file: root.tex
\documentclass[letterpaper, 10 pt, conference]{ieeeconf}  

\IEEEoverridecommandlockouts                              

\usepackage{amsmath, amssymb, amsfonts}
\usepackage{booktabs}
\usepackage{multirow}
\usepackage{graphicx}
\usepackage{balance}

\title{\LARGE \bf
Edge-Cloud Collaborative Reconstruction via Structure-Aware Latent Diffusion for Downstream Remote Sensing Perception
}

\author{Yun Li$^{1}$, Xianju Li$^{1}$%
\thanks{$^{1}$Authors are with the School of Computer Science,
        China University of Geosciences, Wuhan, Hubei, China.
        Emails: {\tt\small 2392717105yun@gmail.com; ddwhlxj@cug.edu.cn}}%
}

\begin{document}

\maketitle
\thispagestyle{empty}
\pagestyle{empty}

\begin{abstract}
The exponential surge in high-resolution remote sensing data faces a severe bottleneck in satellite-to-ground transmission. Limited downlink bandwidth forces the use of extreme high-ratio compression, which irreversibly destroys high-frequency structural details essential for downstream machine perception tasks like object detection. While current super-resolution techniques attempt to recover these details, regression-based methods often yield over-smoothed textures, and generative diffusion models frequently introduce structural hallucinations that mislead detection systems. To address this trade-off, we propose the Structure-Aware Latent Diffusion (SALD) framework, an asymmetric edge-cloud collaborative SR system. At the resource-constrained edge, the system decouples imagery into a highly compressed low-frequency payload and a lightweight soft structural prior. Transmitting this decoupled representation minimizes bandwidth consumption. On the powerful cloud side, we introduce a Structure-Gated Large Kernel (SGLK) module and a Semantic-Guidance Engine (SGE) within the diffusion backbone. These modules leverage the transmitted structural priors to gate large-kernel convolutions, effectively capturing long-range dependencies inherent in aerial scenes while actively suppressing generative hallucinations. Extensive experiments on both the MSCM and UCMerced datasets demonstrate that, even under extreme bandwidth constraints, SALD achieves superior perceptual quality (LPIPS) and significantly enhances downstream performance in both scene classification and small-target detection.
\end{abstract}

\begin{figure*}[t]
    \centering
    \includegraphics[width=\textwidth]{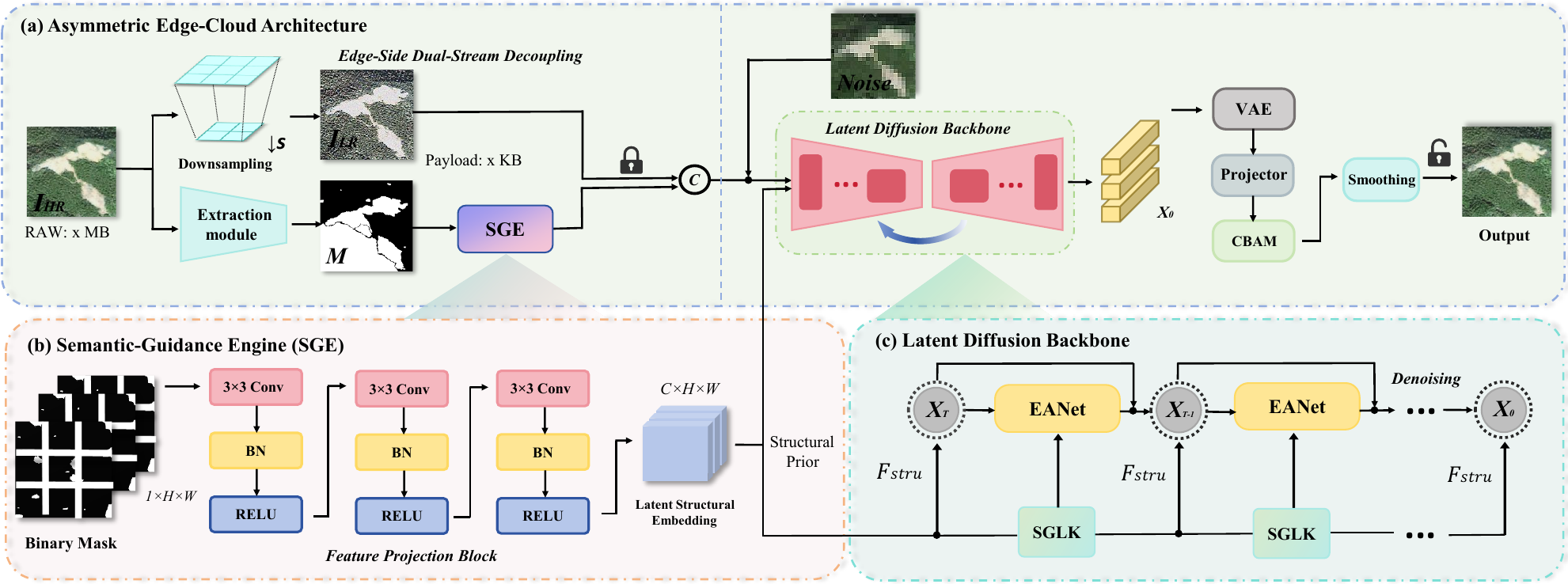}
    \caption{The overall architecture of the proposed Structure-Aware Latent Diffusion (SALD) framework. It adopts an asymmetric edge-cloud collaborative design to achieve bandwidth-efficient super-resolution.}
    \label{fig1}
\end{figure*}

\section{INTRODUCTION}
High-resolution Earth Observation (EO) is critical for applications like maritime surveillance and disaster response. However, growing real-time visual data demands are severely constrained by communication bandwidth limits. In remote environments relying on high-latency satellite links, transmitting raw, petabyte-scale imagery is infeasible. Consequently, systems rely on extreme compression---typically JPEG below 0.1 bits per pixel (bpp)---to maintain operational uplinks.

While aggressive compression reduces payloads, it introduces severe blocking artifacts and erodes high-frequency structural edges. For downstream machine perception, these missing geometric boundaries cause missed small targets or misclassified critical zones, effectively neutralizing the EO data's value.

Image Super-Resolution (SR) techniques aim to recover utility from these degraded bitstreams, but existing paradigms exhibit critical limitations in highly compressed scenarios. Discriminative SR methods minimize pixel-wise errors but predictably yield over-smoothed textures lacking geometric plausibility. Conversely, generative approaches like Latent Diffusion Models (LDM) excel at synthesizing realistic high-frequency textures but are highly susceptible to ``generative hallucinations''. Without strict structural constraints, they may hallucinate debris or distort small-target boundaries, injecting fatal semantic noise.

To address this, we propose the Structure-Aware Latent Diffusion (SALD) framework, reconceptualizing SR as a bandwidth-efficient, edge-cloud collaborative system. In our asymmetric architecture, the resource-limited edge decouples the input into a lightweight low-frequency visual payload and a binary soft structural prior, drastically reducing transmission costs with minimal computation.

At the cloud receiver, we integrate the Semantic-Guidance Engine (SGE) and Structure-Gated Large Kernel (SGLK) module into the diffusion backbone. The SGE maps the discrete prior into a continuous hierarchical latent space. Concurrently, the SGLK utilizes these priors to perform spatially modulated semantic gating over a $9\times9$ depth-wise convolution. Unlike standard $3\times3$ kernels, SGLK efficiently captures long-range dependencies while actively suppressing the hallucinations typical of unconstrained diffusion.

The primary contributions of this work are summarized as follows:
\begin{itemize}
    \item We propose SALD, an asymmetric edge-cloud collaborative SR system that significantly reduces transmission bandwidth by transmitting decoupled structural priors alongside low-resolution payloads.
    \item We design the SGLK and SGE modules, which empower the cloud-side latent diffusion backbone to balance local texture fidelity and global structural integrity, effectively mitigating generative hallucinations.
    \item Extensive experiments validate that SALD achieves state-of-the-art perceptual quality (LPIPS) on both the MSCM and UCMerced datasets \cite{yang2010bag}. Furthermore, on the MSCM dataset, it successfully recovers discriminative features, enabling downstream scene classification and object detection to perform competitively with raw, uncompressed imagery.
\end{itemize}

\section{METHODOLOGY}

\subsection{Problem Formulation}
SALD aims to achieve high-fidelity super-resolution and preserve structural details for downstream machine perception under extreme bandwidth constraints. Unlike methods prioritizing pure compression or pixel-wise distortion, we formulate this as a bandwidth-constrained generative reconstruction problem.

Let $I_{HR} \in \mathbb{R}^{H \times W \times 3}$ denote the high-resolution aerial image acquired at the edge. The satellite-to-ground transmission channel is bounded by a strict bandwidth budget $B_{max}$. Our goal is to learn a lightweight edge encoder $\mathcal{E}$ and a cloud-side generative decoder $\mathcal{D}$ to reconstruct $\hat{I}_{SR}$ such that:
\begin{equation}
    \min_{\mathcal{E}, \mathcal{D}} \mathcal{L}_{total}(I_{HR}, \hat{I}_{SR}) \quad \text{s.t.} \quad \mathcal{R}(\mathcal{E}(I_{HR})) \leq B_{max}
\end{equation}
where $\mathcal{R}(\cdot)$ denotes the bitrate estimation, and $\mathcal{L}_{total}$ evaluates both pixel-wise fidelity and perceptual realism. To solve this, $\mathcal{E}$ decouples the image into a compressed low-frequency visual payload $I_{LR}$ and a lightweight structural prior $M$. The cloud decoder $\mathcal{D}$ leverages these components for structurally-guided latent diffusion.

\begin{figure}[t]
    \centering
    \includegraphics[width=\linewidth]{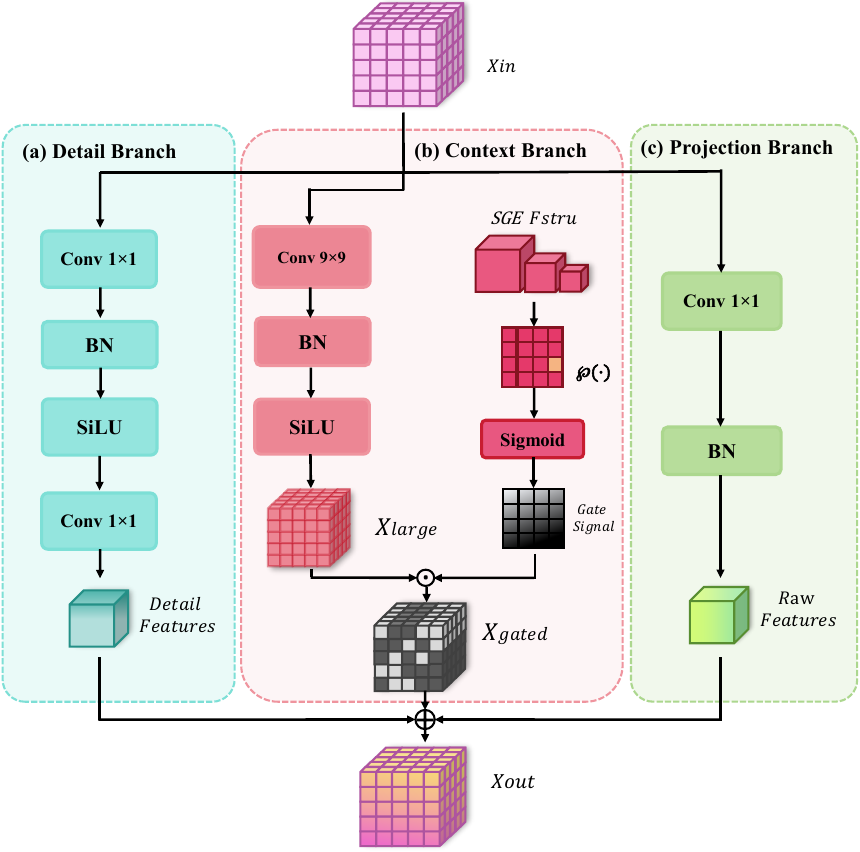}
    \caption{Detailed architecture of the Structure-Gated Large Kernel (SGLK) module. It consists of three parallel branches: a Detail Branch for local feature extraction, a Context Branch that incorporates the structural prior ($\mathcal{P}(\cdot)$) via sigmoid-gated modulation, and a Projection Branch for linear  learning.}
    \label{fig:sglk}
\end{figure}

\subsection{Asymmetric Edge-Cloud Architecture}
As illustrated in Fig.~\ref{fig1}, SALD adopts an asymmetric architecture to balance the computational load between the resource-limited edge and the resource-abundant cloud.

\subsubsection{Edge-side: Payload and Prior Decoupling}
To minimize bandwidth, our Edge-Side Dual-Stream Decoupling strategy downsamples the original high-resolution image $I_{HR}$ by a factor of $s$ ($\downarrow s$) into a highly compressed visual payload $I_{LR} \in \mathbb{R}^{\frac{H}{s} \times \frac{W}{s} \times 3}$. Simultaneously, a lightweight extraction module (saliency detector) extracts a binary semantic mask $M \in \{0, 1\}^{H \times W}$. 

For minimal edge latency, we employ a MobileNetV3-based \cite{mobile} U-Net \cite{ronneberger2015u} ($\approx 1.2$ GFLOPs for $512 \times 512$ inputs) as the extraction module. Crucially, $M$ acts as a soft structural prior rather than a hard filter. If the detector misses a small target, its base visual features remain in the low-frequency payload $I_{LR}$. The cloud strictly utilizes $M$ to enhance high-frequency edges and suppress background hallucinations, ensuring robustness against extraction errors. The tuple $(I_{LR}, M)$ constitutes the final transmitted payload.

\input{table/table1.tex}
\input{table/table2.tex}

\subsubsection{Cloud-side: Structurally-Rectified Restoration}
Upon receiving the payload, the cloud employs a conditional latent diffusion model to reconstruct the high-resolution image. Specifically, the transmitted mask $M$ is first processed by the Semantic-Guidance Engine (SGE) to generate structural prior embeddings. These embeddings are then concatenated (denoted by the C node) with the visual payload $I_{LR}$ and the sampled noise, serving as the joint condition for the Latent Diffusion Backbone.

\input{table/table3.tex}

The backbone iteratively denoises the features to predict the clean latent representation $X_0$. To decode this latent back to the pixel space and refine visual quality, $X_0$ is sequentially passed through a VAE decoder \cite{kingma2013auto}, a Projector, and a Convolutional Block Attention Module (CBAM) \cite{cbam}. Finally, a Smoothing operation is applied to eliminate potential blocky artifacts, yielding the high-fidelity reconstruction output.

\subsection{Preliminaries on Efficient Latent Diffusion}
To leverage diffusion models' generative priors while maintaining cloud deployment efficiency, we build upon the Latent Diffusion Model (LDM) \cite{rombach2022high}, specifically adopting the efficient scheduling and backbone from our previous EDiffSR \cite{ediffsr}. Operating in a compressed latent space, let $X_0 = \mathcal{E}_{vae}(I_{HR})$ be the clean latent representation encoded by a VAE.

\input{table/table4.tex}
\input{table/table5.tex}

The forward process adds Gaussian noise $\epsilon \sim \mathcal{N}(0, \mathbf{I})$ to $X_0$ over $T$ timesteps, producing the noisy latent $X_t = \sqrt{\bar{\alpha}_t} X_0 + \sqrt{1 - \bar{\alpha}_t} \epsilon$. 

As depicted in Fig.~\ref{fig1}c, the reverse denoising process aims to iteratively recover $X_0$ from $X_T$. We train a denoising neural network $\epsilon_\theta(X_t, t, c)$ to predict the added noise. In our architecture, the core denoiser is instantiated as EANet, the fundamental building block in EDiffSR \cite{ediffsr}. Crucially, to effectively guide the generation process, the structural prior $F_{stru}$ and the upsampled payload $I_{LR}^{\uparrow}$ form the joint condition $c$. Rather than a naive concatenation, this condition is systematically processed by our proposed Structure-Gated Large Kernel (SGLK) modules. As illustrated, SGLK extracts multi-scale structural cues to dynamically modulate EANet at each denoising step, ensuring strict structural alignment. The overall learning objective is formulated as:
\begin{equation}
    \mathcal{L}_{diff} = \mathbb{E}_{X_0, t, \epsilon} \left[ \| \epsilon - \epsilon_\theta(X_t, t, c) \|_2^2 \right]
\end{equation}

\subsection{Semantic-Guidance Engine (SGE)}
Integrating discrete binary masks into a continuous latent space causes semantic misalignment. To address this, the SGE module ($\mathcal{F}_{SGE}$) maps the prior $M$ into a hierarchical manifold via a Feature Pyramid Network \cite{lin2017feature}.

\input{table/table6.tex}
\input{table/table7.tex}

As shown in Fig. \ref{fig1}(b), SGE consists of $K=3$ stacked Feature Projection Blocks. Let $F^0 = M$. The transformation at the $k$-th scale ($k \in \{1, ..., K\}$) is:
\begin{equation}
    F^k = \text{ReLU} \left( \mathcal{BN} \left( \text{Conv}_{3\times3} (F^{k-1}) \right) \right)_{\downarrow 2}
\end{equation}
Following feature alignment, a latent structural embedding $\in \mathbb{R}^{C \times H \times W}$ is generated as the structural prior $F_{stru}$.

\subsection{Structure-Gated Large Kernel (SGLK)}
Standard diffusion backbones rely on $3\times3$ convolutions with limited Effective Receptive Fields (ERF) \cite{luo2016understanding}, struggling to capture large-scale topographies. To model long-range dependencies, our SGLK module (Fig. \ref{fig:sglk}) decomposes feature processing into three parallel paths:

1. \textbf{Detail Branch (Fig. \ref{fig:sglk}a):} Employs a $1\times1$ convolution to preserve local channel specificity:
\begin{equation}
X_{detail} = \text{SiLU}(\mathcal{BN}(\text{Conv}_{1\times1}(X_{in}))).
\end{equation}

2. \textbf{Context Branch (Fig.~\ref{fig:sglk}b):} Uses a $9\times9$ convolution, batch normalization, and SiLU activation to aggregate spatial context, producing $X_{large}$. We select $9\times9$ kernels because $3\times3$ kernels fragment structures, while $31\times31$ kernels introduce excessive noise and latency. To incorporate the transmitted prior $F_{stru}$, we apply Semantic Gating instead of heavy conditioning (e.g., Cross-Attention), generating a soft gate $\mathcal{P}(\cdot)$ via Sigmoid activation:
\begin{equation}
   X_{gated} = X_{large} \odot \sigma(\mathcal{P}(F_{stru}))
\end{equation}
where $\odot$ and $\sigma$ denote the Hadamard product and Sigmoid function, respectively. This lightweight design prevents overfitting to degraded masks caused by extraction errors or packet loss. Consequently, the system maintains graceful degradation under severe mask missing rates, utilizing $\mathcal{O}(N)$ complexity to minimize delay.

\input{table/table8.tex}

3. \textbf{Projection Branch (Fig.~\ref{fig:sglk}c):} Instead of a vanilla identity mapping, we employ a linear residual projection $\mathcal{P}_{res}$ to align feature distributions. This branch consists solely of a $1\times1$ convolution followed by batch normalization, avoiding non-linear activations to ensure unimpeded gradient flow and preserve the original latent base. Thus, $X_{res} = \mathcal{P}_{res}(X_{in})$.

Finally, the three branches are aggregated linearly, where the context branch is dynamically modulated by a learnable scale $\gamma$:
\begin{equation}
   X_{out} = X_{detail} + \gamma X_{gated} + X_{res}
\end{equation}

\subsection{Training Objectives}
To solve the optimization problem in Eq.~(1), we unpack $\mathcal{L}_{total}$ into an actionable loss function combining pixel-level fidelity, generative alignment, and perceptual realism:
\begin{equation}
   \mathcal{L}_{total} = \lambda_1 \mathcal{L}_{diff} + \lambda_2 \mathcal{L}_{rec} + \lambda_3 \mathcal{L}_{per}
\end{equation}
where $\mathcal{L}_{diff}$ is the diffusion loss (Eq.~2). To fulfill our formulation, $\mathcal{L}_{rec} = \| I_{HR} - \hat{I}_{SR} \|_1$ strictly enforces low-frequency structural consistency in pixel space to guarantee high-fidelity reconstruction, and $\mathcal{L}_{per} = \sum_{j} \| \phi_j(I_{HR}) - \phi_j(\hat{I}_{SR}) \|_2^2$ computes the feature distance in a pre-trained VGG-19 \cite{johnson2016perceptual} to ensure perceptual realism for downstream tasks. We empirically balance weights at $\lambda_1=1.0$, $\lambda_2=0.1$, and $\lambda_3=0.01$.

\section{EXPERIMENTS}

\subsection{Experimental Setup}

\subsubsection{Implementation Details}
The framework is implemented in PyTorch. For our SALD framework, the diffusion backbone operates with $T=50$ total timesteps. The network is optimized using the AdamW optimizer, where the gradient momentum parameters are set to $\beta_1=0.9$ and $\beta_2=0.999$, alongside a weight decay of $1\times10^{-4}$. To ensure stable convergence, the learning rate follows a cosine annealing schedule, initialized at $2\times10^{-4}$ and gradually decaying to a minimum of $1\times10^{-6}$.

\begin{figure}[t]
    \centering
    \includegraphics[width=\linewidth]{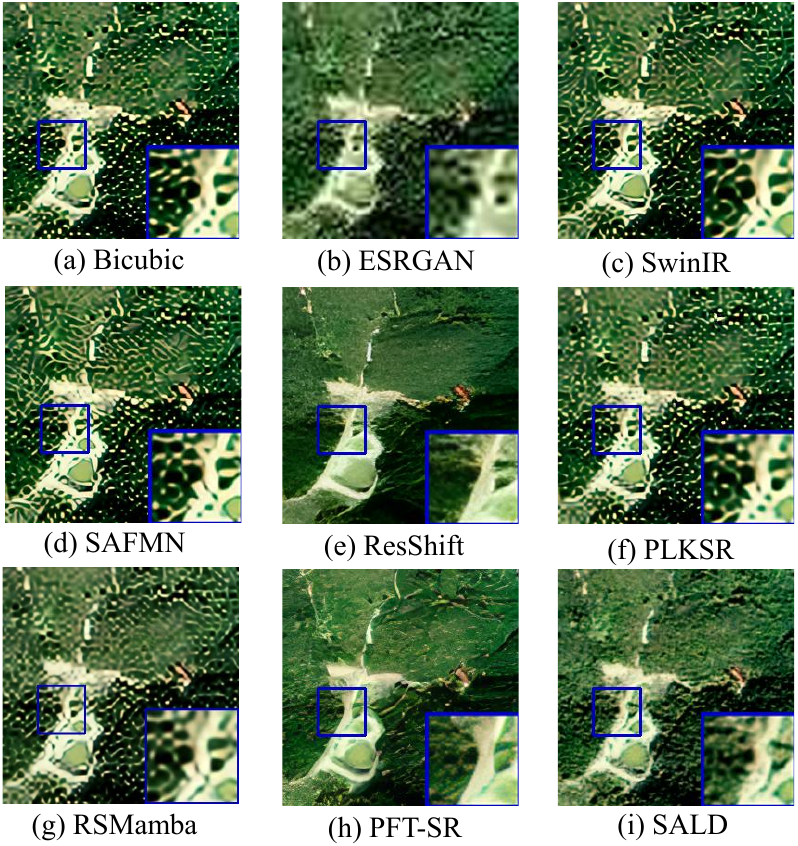} 
    \caption{Visual comparison of different reconstruction methods on the MSCM dataset under extreme bandwidth constraints. SALD accurately preserves critical structural boundaries and restores high-fidelity textures without generative hallucinations.}
    \label{fig3}
\end{figure}

To ensure a rigorously fair comparison, all baseline models are retrained from scratch on our simulated datasets. Rather than applying a uniform optimization strategy, each is trained strictly using its officially recommended settings—including specific optimizers, learning rate schedules, and total epochs—to guarantee optimal convergence for their respective architectures. All methods are evaluated under the identical $0.15$\,MB transmission budget.

The aforementioned experimental protocols and performance evaluations are systematically conducted across two standard remote sensing benchmarks: the MSCM \cite{li2025edg} and UCMerced datasets \cite{yang2010bag}.

\subsubsection{Evaluation Metrics}
Performance is assessed across three dimensions:
(1) Signal Fidelity: PSNR and SSIM \cite{wang2004image};
(2) Generative Realism: LPIPS \cite{lpips} and FID;
(3) Downstream Utility: mAP$_{50}$ and mAP$_{50:95}$ via
YOLOv11 \cite{yolov11} small-vehicle detection.

\subsection{Main Results: Super-Resolution Quality}

As reported in Table~\ref{tab:main_results} and Fig.~\ref{fig3}, SALD achieves
state-of-the-art performance in perceptual metrics (LPIPS and FID)
across both datasets.
Regression-based models (e.g., SwinIR) yield marginally
higher PSNR/SSIM by optimizing for pixel-wise averages, but
fundamentally fail to recover high-frequency structural realism.
Standard diffusion models (e.g., ResShift) improve LPIPS but
suffer from unstable generation.
By leveraging the transmitted structural prior through the SGE and
SGLK modules, SALD bridges this gap, yielding the lowest LPIPS
and FID scores, while maintaining a competitive Params
profile suitable for cloud deployment.

\subsection{Ablation Study}
\label{sec:ablation}

To validate our architectural choices and hyperparameter settings, we perform extensive ablation studies on the MSCM dataset.

\subsubsection{Effectiveness of Core Modules}
To evaluate our proposed components, we conduct a module-wise ablation (Table~\ref{tab2}). The baseline (Model 1), lacking SGE and SGLK, struggles to reconstruct faithful high-frequency details from the compressed payload, yielding the lowest downstream mAP (84.32\%). Introducing either module provides noticeable improvements, but the full SALD framework (Model 4) achieves the best overall performance. This confirms that SGE's structural embeddings and SGLK's dynamic modulation are highly complementary for robust generative guidance.

\subsubsection{Hyperparameter Sensitivity: Sampling Timesteps ($T$)}
A critical bottleneck in latent diffusion is the inference latency dictated by sampling timesteps $T$ \cite{ho2020denoising}. As Table~\ref{tab3} shows, we analyze the trade-off between generative quality and computational cost. Setting $T=50$ emerges as optimal. Increasing $T$ to $100$ or $200$ marginally improves perceptual quality but the marginal gains in downstream perception do not justify the dramatically increased latency. Conversely, aggressive reduction ($T=10$) causes severe structural collapse. Thus, $T=50$ perfectly balances real-time requirements with high-fidelity reconstruction.

\subsubsection{Hyperparameter Sensitivity: Kernel Size ($K$)}
A core Context branch component is the depth-wise convolution kernel utilized for long-range dependencies. Table~\ref{tab4} evaluates perception quality versus computational complexity across kernel sizes ($K \in \{3, 5, 7, 9, 11\}$). Standard $3\times3$ or $5\times5$ kernels lack the receptive field to reconstruct continuous aerial structures, yielding sub-optimal mAP. Increasing $K$ to $11$ does not improve LPIPS but unnecessarily inflates cloud-side parameters, violating efficient deployment objectives. The $9\times9$ kernel optimally balances generative fidelity and system efficiency.

\subsubsection{Hyperparameter Sensitivity: Reconstruction Weight ($\lambda_2$)}
In our joint loss (Eq. 7), $\lambda_2$ governs the perception-distortion trade-off. We fix $\lambda_1=1.0$ and $\lambda_3=0.01$, analyzing $\lambda_2$ in Table~\ref{tab:lambda2_ablation}. A balanced $\lambda_2=0.1$ strikes the optimal equilibrium, achieving the highest visual fidelity (PSNR 23.95) and downstream perception (mAP 87.45\%). Setting $\lambda_2$ too low provides insufficient penalty for structural deviation, allowing the model to generate hallucinations that severely degrade downstream recognition. Conversely, a high weight ($\lambda_2 \geq 0.5$) forces strict pixel-wise minimization, severely suppressing generative diversity. This over-constraint inevitably produces blurred artifacts, deteriorating both LPIPS and mAP.

\subsection{Robustness to Edge Mask Errors}
\label{sec:robustness}

The MobileNetV3-based edge detector cannot guarantee perfect segmentation. To verify resilience against edge mask errors, we simulate miss-detection by zeroing $r\%$ of foreground pixels in $M$, with $r\in\{0, 10, 20, 30, 50\}$.

As shown in Table~\ref{tab6}, SALD exhibits graceful degradation rather than catastrophic collapse. Even at a 30\% miss-detection rate, SALD (mAP@0.5: 85.10) still outperforms the mask-free baseline (84.52) by +0.58 points. This resilience stems from the soft-prior design: missed targets retain their base representations in $I_{LR}$, and the mask functions strictly as an enhancement signal rather than a hard filter. Consequently, partial mask errors degrade high-frequency details but do not eliminate the fundamental structural information in the low-frequency payload.

\subsection{Downstream Machine Perception}
To verify that SALD restores genuine machine utility rather than merely pleasing human perception, we evaluate reconstructions on scene classification and object detection.

\subsubsection{Global Semantics: Scene Classification}
We assess the recovery of global semantics in Table~\ref{tab7}. Discriminative SR networks (e.g., SwinIR, PLKSR) produce over-smoothed textures, limiting robust feature extraction. Unconstrained generative models (e.g., ResShift) perform better but suffer from semantic shifts. SALD surpasses all methods, achieving 90.91\% Top-1 accuracy (outperforming the second-best ResShift by 2.46\%) and closely approaching the uncompressed GroundTruth (92.51\%). This confirms our decoupling effectively preserves the semantic payload under extreme compression.

\subsubsection{Local Geometry: Object Detection}
We deploy YOLOv11 for small vehicle detection, which strictly requires local geometric fidelity (Table~\ref{tab8}). Discriminative networks perform poorly due to blurred edges (causing missed detections), while unconstrained generative models hallucinate fictitious textures (causing false positives). By conditioning the diffusion process on transmitted structural priors, SALD avoids both pitfalls. It achieves the highest mAP@0.5 (87.45\%) and mAP@0.5:0.95 (48.62\%), decisively narrowing the gap to the HR upper bound. This demonstrates SALD safeguards critical boundaries, providing exceptional reliability for collaborative perception systems.

\section{CONCLUSION}
We present SALD, an asymmetric edge-cloud collaborative system designed for bandwidth-constrained remote sensing super-resolution. By decoupling visual payloads and soft structural priors at the edge, and employing the SGE and SGLK modules for structure-gated generation in the cloud, SALD effectively resolves the inherent trade-off between transmission cost, generative hallucination, and downstream machine perception utility. Future work will focus on further quantizing the edge extractor for ultra-low-power satellite hardware.

\balance

\bibliographystyle{IEEEtran}
\bibliography{IEEEabrv, IEEEexample}

\end{document}

%% file: table/table1.tex
\begin{table*}[t]
\caption{Quantitative Comparison on MSCM and UCMerced Datasets}
\label{tab:main_results}
\centering
\small
\renewcommand{\arraystretch}{1.1} 

\begin{tabular*}{\textwidth}{@{\extracolsep{\fill}} l c c cccc cccc @{}}
\toprule
\multirow{2.5}{*}{\hspace{1.5em}\textbf{Method}} & \multirow{2.5}{*}{\textbf{Pub'Year}} & \multirow{2.5}{*}{\textbf{Params (M)}} & \multicolumn{4}{c}{\textbf{MSCM Dataset}} & \multicolumn{4}{c}{\textbf{UCMerced Dataset}} \\
\cmidrule(lr){4-7} \cmidrule(lr){8-11} 
~ & ~ & ~ & PSNR$\uparrow$ & SSIM$\uparrow$ & LPIPS$\downarrow$ & FID$\downarrow$ & PSNR$\uparrow$ & SSIM$\uparrow$ & LPIPS$\downarrow$ & FID$\downarrow$ \\
\midrule 

\hspace{1.5em}Bicubic & - & - & 22.37 & 0.6968 & 0.3604 & 60.34 & 25.21 & 0.7021 & 0.3152 & 59.32 \\
\hspace{1.5em}ESRGAN~\cite{esrgan} & ECCV'18 & 16.88 & 19.89 & 0.5468 & 0.3011 & 80.14 & 24.15 & 0.6857 & 0.2154 & 78.12 \\
\hspace{1.5em}SwinIR~\cite{liang2021swinir} & ICCV'21 & 0.89 & 23.12 & 0.7088 & 0.2945 & 40.32 & 26.42 & 0.7653 & 0.2454 & 35.36 \\
\hspace{1.5em}SAFMN~\cite{safmn} & ICCV'23 & 0.24 & 23.82 & 0.7042 & 0.3128 & 41.12 & 26.98 & 0.7654 & 0.2685 & 43.11 \\
\hspace{1.5em}GRL~\cite{grl} & CVPR'23 & 20.20 & 23.71 & \textbf{0.7395} & 0.2718 & 36.34 & \underline{27.91} & \underline{0.8014} & 0.2295 & 37.81 \\
\hspace{1.5em}ResShift~\cite{resshift} & NeurIPS'23 & 118.59 & 23.82 & 0.6852 & 0.1934 & \underline{32.15} & 25.42 & 0.7126 & \underline{0.1456} & \underline{33.69} \\
\hspace{1.5em}PLKSR~\cite{plksr} & CVPR'24 & 5.32 & \textbf{24.55} & \underline{0.7290} & 0.2801 & 36.12 & 27.65 & 0.7921 & 0.2353 & 34.37 \\
\hspace{1.5em}RSMamba~\cite{rsmamba} & GRSL'24 & 33.1 & 23.94 & 0.7105 & 0.3256 & 35.14 & \textbf{27.95} & \textbf{0.8045} & 0.2284 & 35.24 \\
\hspace{1.5em}PFT-SR~\cite{pft} & CVPR'25 & 19.8 & 23.87 & 0.7126 & \underline{0.1874} & 33.64 & 27.85 & 0.7843 & 0.1892 & 34.65 \\
\hspace{1.5em}InvSR~\cite{invsr} & CVPR'25 & 33.84 & 23.54 & 0.7042 & 0.2136 & 32.95 & 27.56 & 0.7812 & 0.2135 & 34.92 \\
\midrule
\hspace{1.5em}{SALD (Ours)} & - & 27.31 & \underline{23.95} & 0.7127 & \textbf{0.1601} & \textbf{30.21} & 27.89 & 0.7856 & \textbf{0.1105} & \textbf{31.24} \\
\bottomrule
\end{tabular*}
\end{table*}

%% file: table/table2.tex
\begin{table}[htbp]
  \centering
  \caption{Ablation Study of SGE and SGLK Modules on the MSCM Dataset.}
  \label{tab2}
  
  \renewcommand{\arraystretch}{1.1} 

  \resizebox{\linewidth}{!}{%
  \begin{tabular}{@{} c cc ccccc @{}}
    \toprule
    Model & SGE & SGLK & PSNR$\uparrow$ & SSIM$\uparrow$ & LPIPS$\downarrow$ & Top-1(\%) $\uparrow$ & mAP$_{50}(\%)\uparrow$ \\
    \midrule
    1 & $\times$ & $\times$ & 23.32 & 0.7056 & 0.1655 & 86.48 & 84.32 \\
    2 & $\times$ & \checkmark & 23.49 & 0.7083 & 0.1631 & 86.98 & 85.14 \\
    3 & \checkmark & $\times$ & 23.42 & 0.7071 & 0.1618 & 87.91 & 85.43 \\
    4 & \checkmark & \checkmark & \textbf{23.95} & \textbf{0.7127} & \textbf{0.1601} & \textbf{90.91} & \textbf{87.45} \\
    \bottomrule
  \end{tabular}%
  } 
\end{table}

%% file: table/table3.tex
\begin{table}[htbp]
  \centering
  \caption{Hyperparameter sensitivity analysis of the diffusion sampling timesteps ($T$) on the MSCM Dataset.}
  \label{tab3}
  
  \renewcommand{\arraystretch}{1.1}
  
  \resizebox{\linewidth}{!}{%

  \begin{tabular}{c cccccc}
    \toprule
    \textbf{$T$} & PSNR $\uparrow$ & SSIM $\uparrow$ & LPIPS $\downarrow$ & Top-1 (\%) $\uparrow$ & mAP$_{50}$ (\%) $\uparrow$ & Relative Latency \\
    \midrule
    10 & 23.45 & 0.6980 & 0.2315 & 84.20 & 81.20 & 0.2$\times$ \\
    20 & 23.70 & 0.7050 & 0.1980 & 88.15 & 84.50 & 0.4$\times$ \\
    50 & \textbf{23.95} & \textbf{0.7127} & 0.1601 & 90.91 & 87.45 & 1.0$\times$ \\
    100 & \underline{23.92} & \underline{0.7110} & \underline{0.1589} & \underline{90.99} & \underline{87.57} & 2.0$\times$ \\
    200 & 23.88 & 0.7095 & \textbf{0.1585} & \textbf{91.12} & \textbf{87.84} & 4.0$\times$ \\
    \bottomrule
  \end{tabular}%
  }
\end{table}

%% file: table/table4.tex
\begin{table}[htbp]
  \centering
  \caption{Hyperparameter sensitivity analysis of the SGLK kernel size ($K$) on the MSCM Dataset.}
  \label{tab4}
  
  \renewcommand{\arraystretch}{1.1}
  
  \resizebox{\linewidth}{!}{%

  \begin{tabular}{c cccccc}
    \toprule
    \textbf{$K$} & Params (M) & PSNR $\uparrow$ & SSIM $\uparrow$ & LPIPS $\downarrow$ & Top-1 (\%) $\uparrow$ & mAP$_{50}$ (\%) $\uparrow$ \\
    \midrule
    $3\times3$ & 26.85 & 23.65 & 0.7015 & 0.1850 & 86.50 & 84.10 \\
    $5\times5$ & 26.98 & 23.78 & 0.7055 & 0.1760 & 87.80 & 85.60 \\
    $7\times7$ & 27.12 & 23.88 & 0.7085 & 0.1680 & 89.50 & 86.80 \\
    \textbf{$9\times9$} & 27.31 & \textbf{23.95} & \textbf{0.7127} & \textbf{0.1601} & \textbf{90.91} & \textbf{87.45} \\
    $11\times11$ & 27.60 & \underline{23.91} & \underline{0.7114} & \underline{0.1624} & \underline{90.85} & \underline{87.40} \\
    \bottomrule
  \end{tabular}%
  }
\end{table}

%% file: table/table5.tex
\begin{table}[t]
\centering
\caption{Hyperparameter Sensitivity Analysis of the Structural Reconstruction Weight on the MSCM Dataset.}
\label{tab:lambda2_ablation}
\renewcommand{\arraystretch}{1.1}
\setlength{\tabcolsep}{1.8mm}
\begin{tabular}{cccccc}
\toprule
$\lambda_2$ & PSNR$\uparrow$ & SSIM$\uparrow$ & LPIPS$\downarrow$ & Top-1(\%)$\uparrow$ & mAP$_{50}$(\%)$\uparrow$ \\
\midrule
0.01 & 23.21 & 0.6910 & 0.2015 & 84.15 & 82.30 \\
0.05 & 23.75 & 0.7050 & 0.1780 & 88.50 & 85.60 \\
\textbf{0.1} & \textbf{23.95} & \textbf{0.7127} & \textbf{0.1601} & \textbf{90.91} & \textbf{87.45} \\
0.5  & \underline{23.82} & \underline{0.7101} & \underline{0.1720} & \underline{89.20} & \underline{86.10} \\
1.0  & 23.50 & 0.7082 & 0.1950 & 87.50 & 84.20 \\
\bottomrule
\end{tabular}
\end{table}

%% file: table/table6.tex
\begin{table}[htbp]
  \centering
  \caption{Robustness analysis of SALD under varying mask missing rates on the MSCM dataset.}
  \label{tab6}
  
  \renewcommand{\arraystretch}{1.1}
  
  \resizebox{\linewidth}{!}{%

  \begin{tabular}{c ccccc}
    \toprule
    \textbf{Missing Rate (\%)} & PSNR $\uparrow$ & SSIM $\uparrow$ & LPIPS $\downarrow$ & Top-1 (\%) $\uparrow$ & mAP$_{50}$ (\%) $\uparrow$ \\
    \midrule
    0 & \textbf{23.95} & \textbf{0.7127} & \textbf{0.1601} & \textbf{90.91} & \textbf{87.45} \\
    10 & \underline{23.85} & \underline{0.7101} & \underline{0.1610} & \underline{90.15} & \underline{86.80} \\
    20 & 23.74 & 0.7084 & 0.1625 & 89.20 & 85.90 \\
    30 & 23.55 & 0.7004 & 0.1638 & 88.10 & 85.10 \\
    50 & 23.31 & 0.6967 & 0.1645 & 87.05 & 84.60 \\
    \midrule
    w/o Mask & 23.12 & 0.6724 & 0.1659 & 86.78 & 84.52 \\
    \bottomrule
  \end{tabular}%
  }
\end{table}

%% file: table/table7.tex
\begin{table}[h!]
  \centering
  \caption{Scene classification performance on the MSCM dataset.}
  \label{tab7}

  \resizebox{\columnwidth}{!}{
    \begin{tabular}{l ccc c}
      \toprule
      \multirow{2.5}{*}{\hspace{1em}\textbf{Method}} & \multicolumn{3}{c}{\textbf{Accuracy Metrics}} & \multirow{2.5}{*}{\textbf{Avg. Acc.} $\uparrow$} \\
      \cmidrule(lr){2-4}
      ~ & Top-1 (\%) $\uparrow$ & Top-5 (\%) $\uparrow$ & Top-10 (\%) $\uparrow$ & ~ \\
      \midrule
      \hspace{1em}Bicubic     & 79.42 & 85.50 & 93.80 & 86.24 \\
      \hspace{1em}ESRGAN      & 80.15 & 91.50 & 98.10 & 89.92 \\
      \hspace{1em}SwinIR      & 81.25 & 93.50 & 97.50 & 90.75 \\
      \hspace{1em}SAFMN       & 84.85 & 90.40 & 96.20 & 90.48 \\
      \hspace{1em}GRL         & 85.45 & 95.25 & 98.30 & 93.00 \\
      \hspace{1em}ResShift    & \underline{88.45} & \underline{98.20} & \underline{99.85} & \underline{95.50} \\
      \hspace{1em}PLKSR       & 86.24 & 97.90 & 98.40 & 94.18 \\
      \hspace{1em}RSMamba     & 86.50 & 94.80 & 98.45 & 93.25 \\
      \hspace{1em}PFT-SR      & 88.33 & 97.65 & 98.46 & 94.81 \\
      \hspace{1em}InvSR       & 87.12 & 95.80 & 98.45 & 93.79 \\
      \hspace{1em}\textbf{SALD} & \textbf{90.91} & \textbf{99.15} & \textbf{99.92} & \textbf{96.66} \\
      \midrule
      \hspace{1em}{GroundTruth} & 92.51 & 99.41 & 99.95 & 97.29 \\
      \bottomrule
    \end{tabular}
  }
\end{table}

%% file: table/table8.tex
\begin{table}[h!]
  \centering
  \caption{Object detection performance on the MSCM dataset.}
  \label{tab8}
  
  \small 
  \setlength{\tabcolsep}{12pt}

  \begin{tabular}{l cc}
    \toprule
    \hspace{1em}\textbf{Method} & mAP$_{50}$ (\%) $\uparrow$ & mAP$_{50:95}$ (\%) $\uparrow$ \\
    \midrule
    \hspace{1em}Bicubic & 61.25 & 29.85 \\
    \hspace{1em}ESRGAN & 69.65 & 31.20 \\
    \hspace{1em}SwinIR & 71.45 & 37.25 \\
    \hspace{1em}SAFMN & 72.95 & 38.61 \\
    \hspace{1em}GRL & 78.92 & 40.12 \\
    \hspace{1em}ResShift & \underline{84.82} & \underline{46.15} \\
    \hspace{1em}PLKSR & 75.68 & 39.90 \\
    \hspace{1em}RSMamba & 77.88 & 40.80 \\
    \hspace{1em}PFT-SR & 84.35 & 45.62 \\
    \hspace{1em}InvSR  & 83.41 & 44.65 \\
    \hspace{1em}\textbf{SALD} & \textbf{87.45} & \textbf{48.62} \\
    \midrule 
    \hspace{1em}GroundTruth & 89.51 & 50.23 \\
    \bottomrule
  \end{tabular}
\end{table}